\documentclass[10pt,letterpaper]{article}

\usepackage{cogsci}
\usepackage{pslatex}
\usepackage{apacite}

\usepackage{natbib}
\usepackage[dvipsnames]{xcolor}
\usepackage{graphicx}
\graphicspath{{./figs/}}
\usepackage{wrapfig}
\usepackage{amsmath}
\usepackage{float}

\title{Capturing human category representations by sampling in deep feature spaces}
 
\author{{\bf Joshua C. Peterson\textsuperscript{\scriptsize 1} (jpeterson@berkeley.edu)} \\
    {\bf Jordan W. Suchow\textsuperscript{\scriptsize 1} (suchow@berkeley.edu)} \\
    {\bf Krisha Aghi\textsuperscript{\scriptsize 2} (kaghi@berkeley.edu)} \\
    {\bf Alexander Y. Ku\textsuperscript{\scriptsize 1} (alexku@berkeley.edu)} \\
    {\bf Thomas L. Griffiths\textsuperscript{\scriptsize 1} (tom\_griffiths@berkeley.edu)} \\
  Department of Psychology\textsuperscript{\scriptsize 1}, Helen Wills Neuroscience Institute\textsuperscript{\scriptsize 2} \\ University of California, Berkeley, CA 94720 USA}

\begin{document}

\maketitle

\begin{abstract}

Understanding how people represent categories is a core problem in cognitive science. Decades of research have yielded a variety of formal theories of categories, but validating them with naturalistic stimuli is difficult. The challenge is that human category representations cannot be directly observed and running informative experiments with naturalistic stimuli such as images requires a workable representation of these stimuli. Deep neural networks have recently been successful in solving a range of computer vision tasks and provide a way to compactly represent image features. Here, we introduce a method to estimate the structure of human categories that combines ideas from cognitive science and machine learning, blending human-based algorithms with state-of-the-art deep image generators. We provide qualitative and quantitative results as a proof-of-concept for the method's feasibility. Samples drawn from human distributions rival those from state-of-the-art generative models in quality and outperform alternative methods for estimating the structure of human categories.


\textbf{Keywords:} 
categorization; neural networks; Markov Chain Monte Carlo
\end{abstract}

\section{Introduction}

Categorization is a central problem in cognitive science and concerns why and how we divide the world into discrete units at various levels of abstraction. The biggest challenge for studying human categorization is that the content of mental category representations cannot be directly observed, which has led to development of laboratory methods for estimating this content from human behavior. Because these methods rely on small sets of artificial stimuli with handcrafted or low-dimensional feature sets, they are ill-suited to the study of categorization as an intelligent process, which is principally motivated by robust human categorization performance in complex ecological settings \citep{nosofsky2017toward}.

One of the challenges of applying laboratory methods to realistic stimuli such as natural images is finding a way to represent them. Deep learning models, such as convolutional neural networks, discover features that can be used to represent complex images compactly and perform well on a range of computer vision tasks \citep{lecun2015deep}. It may be possible to express human category structure using these features, an idea supported by recent work in cognitive science \citep{lake2015deep,peterson2016adapting}.

Ideally, experimental methods could be combined with state-of-the-art deep learning models to estimate the structure of human categories with as few assumptions as possible, and while avoiding the problem of dataset bias. In what follows, we propose a method that uses a human in the loop to estimate arbitrary distributions over complex feature spaces, adapting an existing experimental paradigm to exploit advances in deep architectures to capture the precise structure of human category representations and iteratively sharpen them. Such knowledge is crucial to forming an ecological theory of intelligent categorization behavior and to providing a ground-truth benchmark to guide future work in machine learning.

\section{Background}

\subsubsection{Deep neural networks for images}
Deep neural networks are modern instantiations of classic multilayer perceptrons, and represent a powerful class of machine learning model. DNNs can be trained efficiently through gradient descent and structurally specialized for particular domains \citep{lecun2015deep}. In the image domain, deep convolutional neural networks (CNNs; \citealp{lecun1989backpropagation}) excel in classic computer vision tasks, including natural image classification \citep{krizhevsky2012imagenet}. CNNs exploit knowledge of the input domain by learning a hierarchical set of translation-invariant image filters. The resulting representations, real-valued feature vectors, are surprisingly general and outperform other methods in explaining complex human behavior \citep{lake2015deep,peterson2016adapting}.


Generative Adversarial Networks (GANs; \citealp{goodfellow2014}) and Variational Autoencoders (VAEs; \citealp{wellingvae2013}) provide a generative approach to modeling the content of natural images. Importantly, though the approaches differ considerably, each approach makes use of a network (called a ``decoder'' or ``generator'') that learns a deterministic function that maps samples from a known noise distribution $p(z)$ (e.g., a multivariate Gaussian) to samples from the true image distribution $p(x)$. This can be thought of as mapping a relatively low-dimensional feature representation $z$ to a relatively high-dimensional image $x$. Sampling new images from these networks is as simple as passing Gaussian noise into the learned decoder. In addition, because of its simple form, the resulting latent space $z$ tends to be easy to traverse meaningfully (i.e., an intrinsic linear manifold) and can be readily visualized via the decoder, a property we exploit presently.


\begin{figure*}
  \begin{minipage}[c]{0.53\textwidth}
    \includegraphics[width=\textwidth]{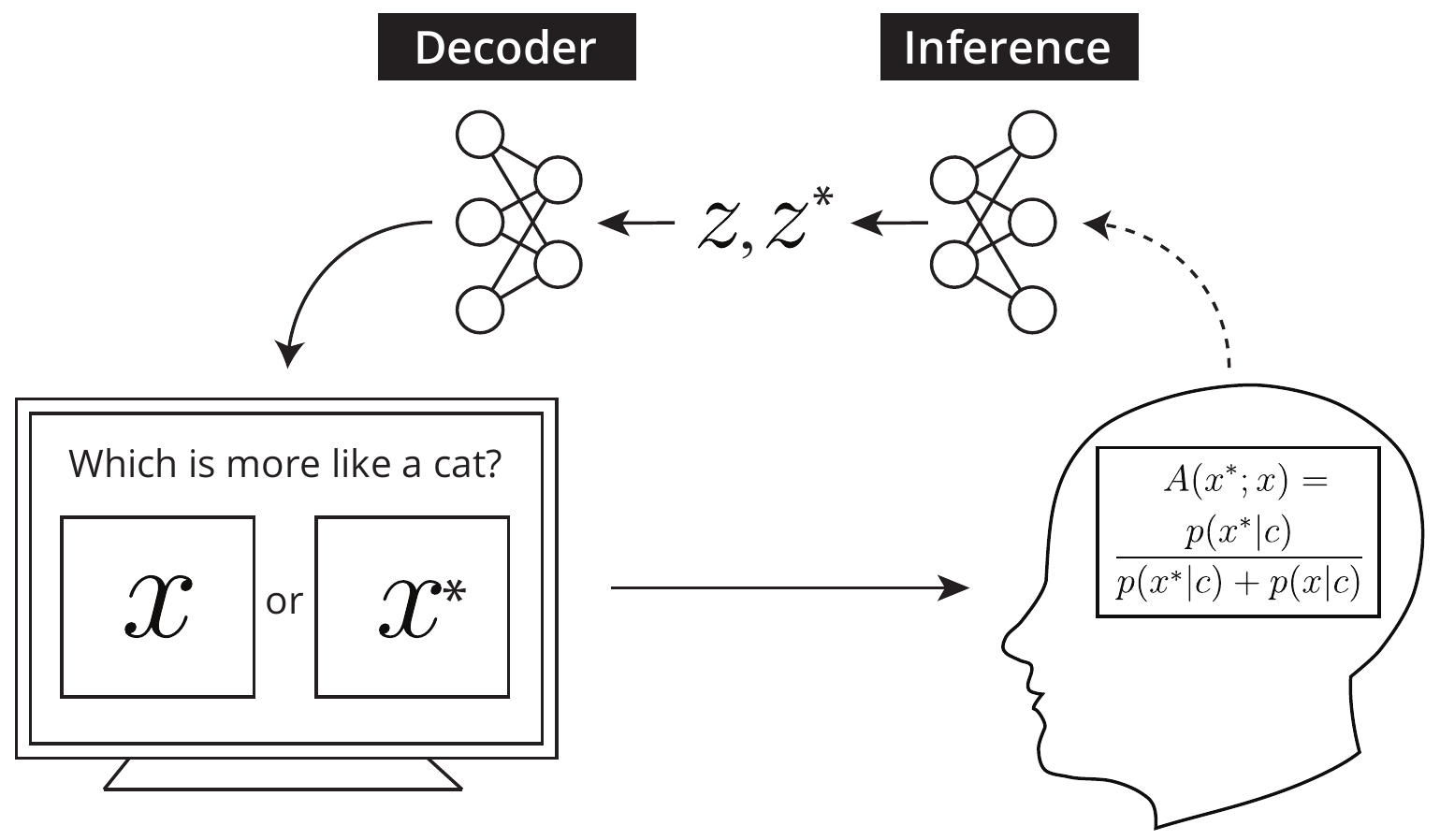}
  \end{minipage}\hfill
  \begin{minipage}[c]{0.45\textwidth}
    \caption{Deep MCMCP. A current state $z$ and proposal $z^*$ (top middle) are fed to a pretrained deep image generator/decoder network (top left). The corresponding decoded images $x$ and $x^*$ for the two states are presented to human raters on a computer screen (leftmost arrow and bottom left). Human raters then view the images in an experiment (bottom middle arrow) and act as part of an MCMC sampling loop, choosing between the two states/images in accordance with the Barker acceptance function (bottom right). The chosen image can then be sent to the inference network (rightmost arrow) and decoded in order to select the state for the next trial, however this step is unnecessary when we know exactly which states corresponds to which images.}
    \label{method}
  \end{minipage}
  \vspace{-3mm}
\end{figure*}

\subsubsection{Estimating the structure of human categories}

Methods for estimating human category templates have existed for some time. In psychophysics, the most popular and well-understood method is known as \textit{classification images} (CI; \citealp{ahumada1996perceptual}).

In the classification images experimental procedure, a human participant is presented with images from two categories, A and B, each with white noise overlaid, and asked to select the stimulus that corresponds to the category in question. On most trials, the participant will obviously select the exemplar generated from the category in question. However, if the added white noise significantly perturbs features of the image that are important to making the distinction, they may fail. Exploiting this, we can estimate the decision boundary from a number of these trials using the simple formula:
\begin{equation}
    (n_{AA} + n_{BA}) - (n_{AB} + n_{BB}),
\end{equation}
where $n_{XY}$ is the average of the noise across trials where the correct class is $X$ and the observer chooses $Y$.

\cite{vondrick2015learning} used a variation on classification images using deep image representations that could be inverted back to images using an external algorithm. In order to avoid dataset bias introduced by perturbing real class exemplars, white noise in the feature space was used to generate stimuli. In this special case, category templates reduce to $n_{A}-n_{B}$. On each trial of the experiment, participants were asked to select which of two images (inverted from feature noise) most resembled a particular category. Because the feature vectors for all trials were random, thousands of stimuli could be rendered in advance of the experiment using relatively slow methods that require access to large datasets. This early inversion method was applied to mean feature vectors for thousands of positive choices in the experiments and yielded qualitatively decipherable category template images, as well as better objective classification decision boundaries that were guided human bias. Under the assumption that human category distributions are Gaussian with equal variance, this method yields a vector that aligns with the nearest-mean decision boundary, although a massive number of human trials are required.


Markov Chain Monte Carlo with People (MCMCP; \citealp*{sanborn2007markov}), an alternative to classification images, is an experimental procedure in which humans act as a valid acceptance function $A$ in the Metropolis--Hastings algorithm, exploiting the fact that Luce's choice axiom, a well-known model of human choice behavior, is equivalent to the Barker acceptance function (see equation in Figure \ref{method}). On the first trial, a stimulus $x$ is drawn arbitrarily from the parameter space and compared to a new proposed stimulus $x^*$ that is nearby in that parameter space. The participant makes a forced choice as to which is the better exemplar of some category (e.g., dog), acting as the acceptance function $A(x^*; x)$. If the initial stimulus is chosen, the Markov chain remains in that state. If the proposed stimulus is chosen, the chain moves to the proposed state. The process then repeats until the chain converges to the target category distribution $p(x|c)$. In practice, convergence is assessed heuristically, or limited by the number of human trials that can be practically obtained. 

MCMCP has been successfully employed to capture a number of different mental categories \citep{sanborn2007markov,martin2012testing}, and though these spaces are higher-dimensional than those in previous laboratory experiments, they are still relatively small and artificial compared to real images. Unlike classification images, this method makes no assumptions about the structure of the category distributions and thus can estimate means, variances, and higher order moments. Therefore, we take it as a starting point for the current method.


\section{MCMCP in deep feature spaces}

The typical MCMCP experiment is effective so long as noise can be added to dimensions in the stimulus parameter space to create meaningful changes in content. In the case of natural images, noise in the space of all pixel intensities is very unlikely to modify the stimulus in meaningful ways. Instead, we propose perturbing images in a deep feature space that captures only essential variation. Since trials in an MCMCP experiment are not independent, we employ real-time, web-accessible generative adversarial networks to render high quality inversions from their latent features. The mapping from features to images learned by a GAN is deterministic, and therefore MCMCP in low-dimensional feature space approximates the same process in high-dimensional image space. The resulting judgments (samples) approximate distributions that both derive arbitrary human category boundaries for natural images and can be sampled from to create images, yielding new human-like generative image models. A schematic of this procedure is illustrated in Figure \ref{method}.

There are several theoretical advantages to our method over previous efforts. First, MCMCP can capture arbitrary distributions, so it is not as sensitive to the structure of the underlying low-dimensional feature space and should provide better category boundaries than classification images when required. This is important when using various deep features spaces that were learned with different training objectives and architectures. MCMC inherently spends less time in low probability regions and should in theory waste fewer trials. Having generated the images online and as a function of the participant's decisions, there is no dataset or sampling bias, and auto-correlation can be addressed by removing temporally adjacent samples from the chain. Finally, using a deep generator provides drastically clearer samples than shallow reconstruction methods, and can be trained end-to-end with an inference network that allows us to categorize new images using the learned distribution.

\section{Experiments}

For our experiments, we explored two image generator networks trained on various datasets. Since even relatively low-dimensional deep image embeddings are large compared to controlled laboratory stimulus parameter spaces, we use a hybrid proposal distribution in which a Gaussian with a low variance is used with probability $P$ and a Gaussian with a high variance is used with probability $1-P$. This allows participants to both refine and escape nearby modes, but is simple enough to avoid excessive experimental piloting that more advanced proposal methods often require.\

Participants in all experiments completed exactly 64 trials (image comparisons), collectively taking about $5$ minutes, containing segments of several chains for multiple categories. The order of the categories and chains within those categories were always interleaved. Each participant's set of chains for each category were initialized with the previous participants final states, resulting in large, multi-participant chains. All experiments were conducted on Amazon Mechanical Turk. If a single image did not load for a single trial, the data for the subject undergoing that trial was completely discarded, and a new subject was recruited to continue on from the original chain state.

\begin{figure*}[!ht]
  \begin{center}
    \includegraphics[width=0.8\textwidth]{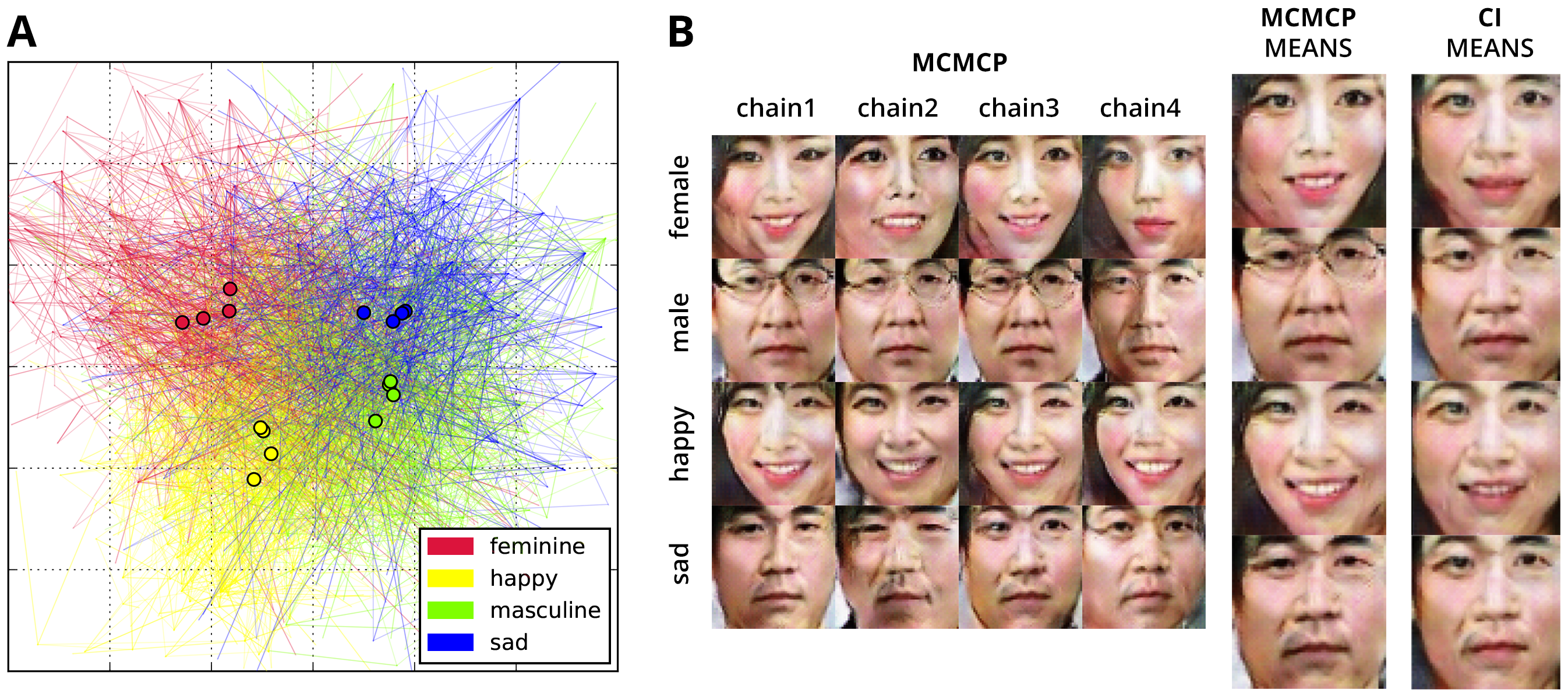}
  \vspace{-2mm}
  \caption{Visualizing captured representations. {\bf A.} Fisher Linear Discriminant projections of all four MCMCP chains for each of the four face categories. The four sets of chains overlap to some degree, but are also well-separated overall. Means of individual chains are closer to other means from the same class than to those of other classes. {\bf B.} Individual MCMCP chain means ($4\times4$ grid) and overall category means (second to last) visualized as images (overall CI means also shown for comparison in the final column).}
  \label{face-chains-and-means}
  \end{center}
\vspace{-5mm}
\end{figure*}

\begin{figure}[!b]
  \begin{center}
    \includegraphics[trim = 5mm 5mm 13mm 12mm, clip, width=1.0\linewidth]{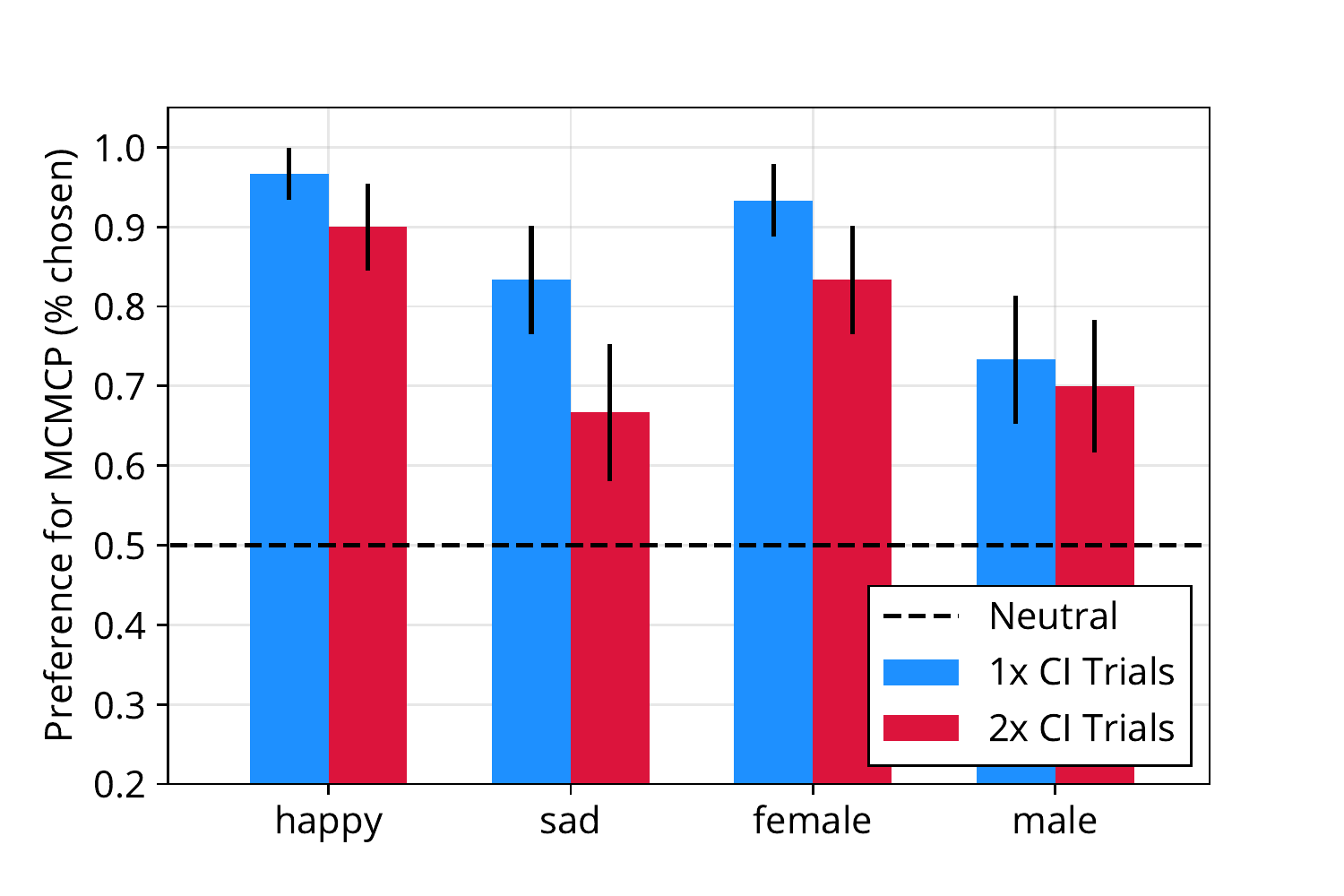}
  \vspace{-7mm}
  \caption{Human two-alternative forced-choice tasks reveal a strong preference for MCMCP means as representations of a category, when twice as many trials are used for CI.}
  \label{faces_mcmcp_vs_ci}
  \end{center}
\end{figure}


\subsection{Experiment 1: Initial test with face categories}
\subsubsection{Methods}
We first test our method using DCGAN \citep{radford2015unsupervised} trained on the Asian Faces Dataset. We chose this dataset because it requires a deep architecture to produce reasonable samples (unlike MNIST, for example), yet it is constrained enough to test-drive our method using a relatively simple latent space. Four chains for each of four categories (male, female, happy, and sad) were used. Proposals were generated from an isometric Gaussian with a standard deviation of 0.25 50\% of the time, and $2$ otherwise. In addition, we conducted a baseline in which two new initial state proposals were drawn on every trial, and were independent of previous trials (classification images). The final dataset contained $50$ participants and over $3,200$ trials (samples) in total for all chains. The baseline classification images (CI) dataset contained the same number of trials and participants.

\begin{figure*}[!ht]
  \begin{center}
    \includegraphics[width=0.82\textwidth]{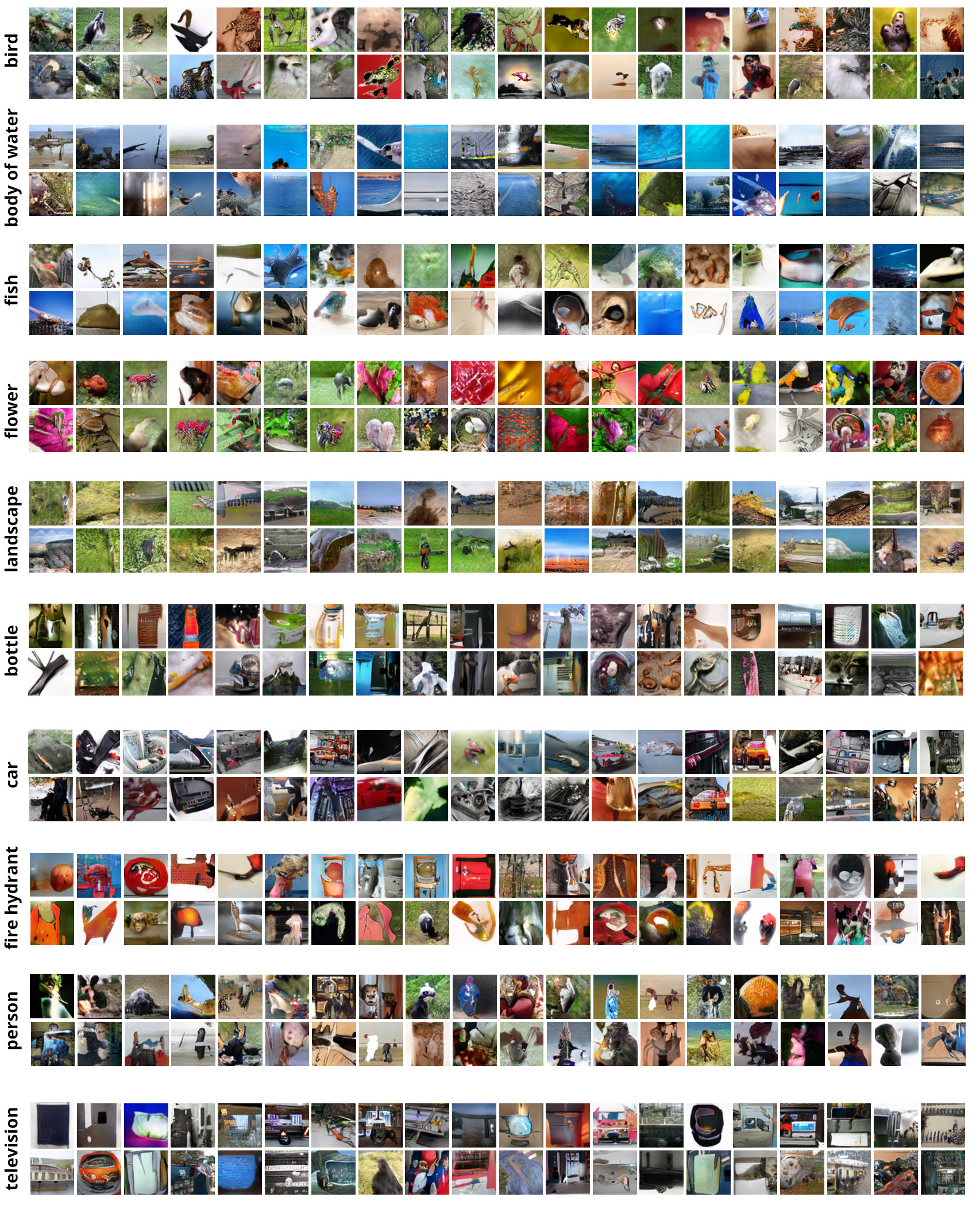}
  \vspace{-5mm}
  \caption{40 most interpretable mixture component means (modes) taken from the 50 largest mixture weights for category.}
  \label{samples}
  \end{center}
\vspace{-5mm}
\end{figure*}

\subsubsection{Results}
MCMCP chains are visualized using Fisher Linear Discriminant Analysis in Figure \ref{face-chains-and-means}, along with the resulting averages for each chain and each category. Chain means within a category show interesting variation, yet converge to similar regions in the latent space as expected. Figure \ref{face-chains-and-means} also shows visualizations of the mean faces for both methods in the final two columns. MCMCP means appear to have converged quickly, whereas CI means only moderately resemble their corresponding category (e.g., the MCMCP mean for ``happy'' is fully smiling, while the CI mean barely reveals teeth). All four CI means appear closer to a mean face, which is what one would expect from averages of noise. We validated this improvement with a human experiment in which $30$ participants made forced choices between CI and MCMCP means. The results are reported in Figure \ref{faces_mcmcp_vs_ci}. MCMCP means are consistently highly preferred as representations of each category as compared to CI. This remained true even when an additional $50$ participants (total of $100$) completed the CI task, obtaining twice as many image comparison trials as with MCMCP.


\subsection{Experiment 2: Larger networks \& larger spaces}

The results of Experiment 1 show that reasonable category templates can be obtained using our method, yet the complexity of the stimulus space used does not rival that of large object classification networks. In Experiment 2, we tackled a more challenging (and interesting) form of the problem. To do this, we employed a bidirectional generative adversarial network (BiGAN; \citealp{donahue2016adversarial}) trained on the 1.2 million-image \textsc{ilsvrc12} dataset ($64\times64$ center-cropped). BiGAN includes an inference network, which regularizes the rest of the model and produces unconditional samples competitive with the state-of-the-art. This also allows for the later possibility of comparing human distributions with other networks as well as assessing machine classification performance with new images based on the granular human biases captured.

\subsubsection{Methods}
Our generator network was trained given uniform rather than Gaussian noise, which allows us to avoid proposing highly improbable stimuli to participants. Additionally, we avoid proposing states outside of this hypercube by forcing $z$ to wrap around (proposals that travel outside of $z$ are injected back in from the opposite direction by the amount originally exceeded). In particular, we run our MCMC chains through an unbounded state space by redefining each bounded dimension $z_k$ as
\newcommand\floor[1]{\lfloor#1\rfloor}
\vspace{-1mm}
\begin{equation}
    z^\prime_k = 
\begin{cases}
  -sgn(z_k) \times [1-(z_k-\floor{z_k})], & \text{if } {\mid}z{\mid} > 1 \\
  z_k, & \text{otherwise}.
\end{cases}
\vspace{-1mm}
\end{equation}
Proposals were generated from an isometric Gaussian with a standard deviation of $0.1$ 60\% of the time, and $0.7$ otherwise. 
%

We use this network to obtain large chains for two groups of five categories. Group 1 included \textit{bottle}, \textit{car}, \textit{fire hydrant}, and \textit{person}, \textit{television}, following \cite{vondrick2015learning}. Group 2 included  \textit{bird}, \textit{body of water}, \textit{fish}, \textit{flower}, and \textit{landscape}. Each chain was approximately $1,040$ states long, and four of these chains were used for each category (approximately $4,160$). In total, across both groups of categories, we obtained exactly $41,600$ samples from $650$ participants. 

To demonstrate the efficiency and flexibility of our method compared to alternatives, we obtained an equivalent number of trials for all categories using the variant of classification images introduced in \cite{vondrick2015learning}, with the exception that we used our BiGAN generator instead of the offline inversion previously used. This also serves as an important baseline against which to quantitatively evaluate our method because it estimates the simplest possible template.

\subsubsection{Results}
\begin{figure*}[!ht]
  \begin{center}
    \includegraphics[trim = 0mm 0mm 0mm 0mm, clip, width=1.0\textwidth]{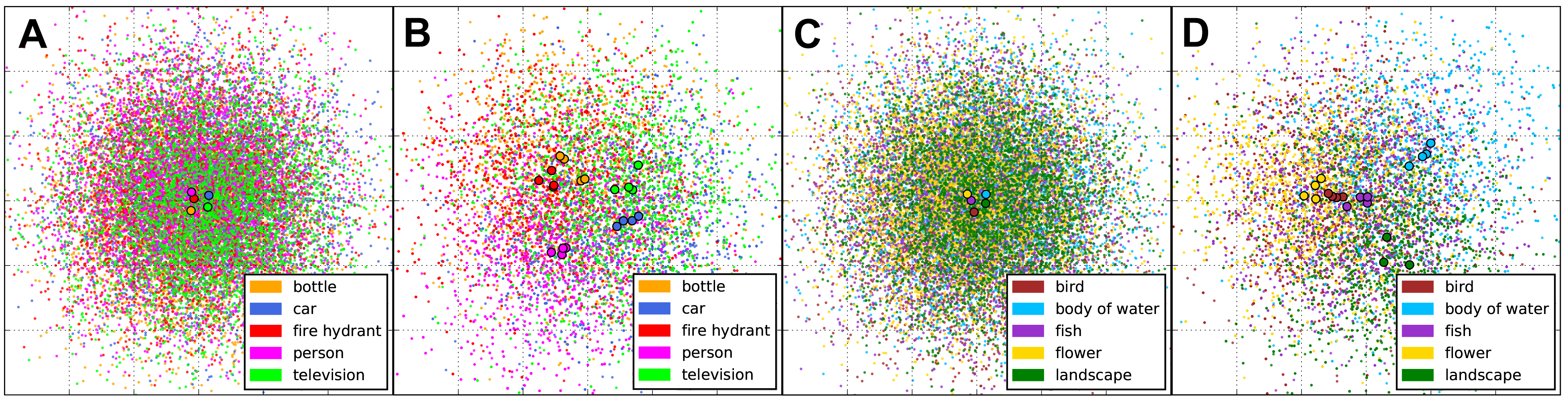}
  \vspace{-6mm}
  \caption{Categories are better separated by MCMCP representations. Fisher Linear Discriminant projections of \textbf{A}. CI comparisons for each category of group 1, \textbf{B.} samples for MCMCP chains for category group 1, \textbf{C.} CI comparisons for each category of group 2, and \textbf{D.} samples for MCMCP chains for category group 2. For A and C, large dots represent category means. For B and D, large dots represent chain means.}
  \label{bigan-results}
  \end{center}
\vspace{-5mm}
\end{figure*}

The acceptance rate was approximately $50\%$ for both category groups, which is near the common goal for MCMCP experiments. The samples for all ten categories are shown in Figure \ref{bigan-results}B and D using Fisher Linear Discriminant Analysis. Similar to the face chains, the four chains for each category converge to similar regions in space, largely away from other categories. In contrast, classification images shows little separation with so few trials (\ref{bigan-results}C and D). Previous work suggests that at least an order of magnitude higher number of comparisons may be needed for satisfactory estimation of category means. Our method estimates well-separated category means in a manageable number of trials, allowing for the method to scale greatly. This makes sense given that CI compares arbitrary images, potentially wasting many trials, and clearly suffers from a great deal of noise.

Beyond yielding a decision rule, our method additionally produces a density estimate of the entire category distribution. In classification images, only mean template images can be viewed, while we are able to visualize several modes in the category distribution. Figure \ref{samples} visualizes these modes using the means of each component in a mixture of Gaussians density estimate. This produces realistic-looking multi-modal mental category templates, which to our knowledge has never been accomplished with respect to natural image categories.

\subsection{Efficacy in classifying real images}

Improvements of MCMCP over classification images may be both perceptible and detectable, but their practical differences are also worth considering --- do they differ significantly on real-world tasks? Moreover, if the representations we learn through MCMCP are good approximations to people, we would expect them to perform reasonably well in categorizing real images. For this reason, we provide an additional quantitative assessment of the samples we obtained and compare them to classification images (CI) using an external classification task. 

To do this, we scraped $\approx500$ images from Flickr for each of the ten categories, which was used for a classification task. To classify the images using our human-derived samples, we used (1) the nearest-mean decision rule, and (2) a decision rule based on the highest log-probability given by our ten density estimates. For classification images, only a nearest-mean decision rule can be tested. In all cases, decision rules based on our MCMCP-obtained samples overall outperform a nearest-mean decision rule using classification images (see Table \ref{class-results}). In category group 1, the MCMCP density performed best and was more even across classes. In category group 2, nearest-mean using our MCMCP samples did much better than a density estimate or CI-based nearest-mean.

\begin{table}
\caption{Classification performance compared to chance for both category sets (chance is 0.20).}
\begin{center}
\setlength\tabcolsep{3.1pt}
\begin{tabular}{ lcccccc } 
 \hline
 & \textbf{bird} & \textbf{body of water} & \textbf{fish} & \textbf{flower} & \textbf{landscape} & \textbf{all}\\ 
 \hline
 MM & .33 & .28 & .01 & .57 & .67 & .37 \\ 
 MD & .23 & .31 & .18 & .44 & .73 & \textbf{.38} \\ 
 CM & .23 & .30 & .2 & .24 & .52 & .30 \\ 
 \hline
 & \textbf{bottle} & \textbf{fire hydrant} & \textbf{car} & \textbf{person} & \textbf{television} & \textbf{all}\\ 
 \hline
 MM & .15 & .11 & .32 & .77 & .73 & \textbf{.42} \\ 
 MD & .25 & .26 & .56 & .19 & .50 & .35 \\ 
 CM & .28 & .15 & .62 & .12 & .13 & .26 \\ 
 \hline
\end{tabular}
\end{center}
\vspace{-1mm}
\small{MM = MCMCP Mean, MD = MCMCP Density, CM = CI Mean}
\label{class-results}
\vspace{-2mm}
\end{table}

\section{Discussion}

Our results demonstrate the potential of our method, which leverages both psychological methods and deep surrogate representations to make the problem of capturing human category representations tractable. The flexibility of our method in fitting arbitrary generative models allows us to visualize multi-modal category templates for the first time, and improve on human-based classification performance benchmarks. It is difficult to guarantee that our chains explored enough of the relevant space to actually capture the concepts in their entirety, but the diversity in the modes visualized and the improvement in class separation achieved are positive indications that we are on the right track. Further, the framework we present can be straightforwardly improved as generative image models advance, and a number of known methods for improving the speed, reach, and accuracy of MCMC algorithms can be applied to MCMCP make better use of costly human trials.

There are several obvious limitations of our method. First, the structure of the underlying feature spaces used may either lack the expressiveness (some features may be missing) or the constraints (too many irrelevant features or possible images wastes too many trials) needed to map all characteristics of human mental categories in a practical number of trials. Even well-behaved spaces are very large and require many trials to reach convergence. Addressing this will require continuing exploration of a variety of generative image models. We see our work as part of an iterative refinement process that can yield more granular human observations and inform new deep network objectives and architectures, both of which may yet converge on a proper, yet tractable model of real-world human categorization.

\renewcommand{\bibliographytypesize}{\small}
\bibliographystyle{apacite}
\setlength{\bibleftmargin}{.125in}
\setlength{\bibindent}{-\bibleftmargin}
\bibliography{citations}

\begin{thebibliography}{}

\bibitem [\protect \citeauthoryear {%
Ahumada%
}{%
Ahumada%
}{%
{\protect \APACyear {1996}}%
}]{%
ahumada1996perceptual}
\APACinsertmetastar {%
ahumada1996perceptual}%
\begin{APACrefauthors}%
Ahumada.%
\end{APACrefauthors}%
\unskip\
\newblock
\APACrefYearMonthDay{1996}{}{}.
\newblock
{\BBOQ}\APACrefatitle {Perceptual classification images from Vernier acuity
  masked by noise} {Perceptual classification images from vernier acuity masked
  by noise}.{\BBCQ}
\newblock
\APACjournalVolNumPages{Perception}{25}{}{2--2}.
\PrintBackRefs{\CurrentBib}

\bibitem [\protect \citeauthoryear {%
Donahue%
, Kr{\"a}henb{\"u}hl%
\BCBL {}\ \BBA {} Darrell%
}{%
Donahue%
\ \protect \BOthers {.}}{%
{\protect \APACyear {2016}}%
}]{%
donahue2016adversarial}
\APACinsertmetastar {%
donahue2016adversarial}%
\begin{APACrefauthors}%
Donahue, J.%
, Kr{\"a}henb{\"u}hl, P.%
\BCBL {}\ \BBA {} Darrell, T.%
\end{APACrefauthors}%
\unskip\
\newblock
\APACrefYearMonthDay{2016}{}{}.
\newblock
{\BBOQ}\APACrefatitle {Adversarial feature learning} {Adversarial feature
  learning}.{\BBCQ}
\newblock
\APACjournalVolNumPages{arXiv preprint arXiv:1605.09782}{}{}{}.
\PrintBackRefs{\CurrentBib}

\bibitem [\protect \citeauthoryear {%
Goodfellow%
\ \protect \BOthers {.}}{%
Goodfellow%
\ \protect \BOthers {.}}{%
{\protect \APACyear {2014}}%
}]{%
goodfellow2014}
\APACinsertmetastar {%
goodfellow2014}%
\begin{APACrefauthors}%
Goodfellow, I.%
, Pouget-Abadie, J.%
, Mirza, M.%
, Xu, B.%
, Warde-Farley, D.%
, Ozair, S.%
\BDBL {}Bengio, Y.%
\end{APACrefauthors}%
\unskip\
\newblock
\APACrefYearMonthDay{2014}{}{}.
\newblock
{\BBOQ}\APACrefatitle {Generative Adverserial Networks} {Generative adverserial
  networks}.{\BBCQ}
\newblock
\BIn{} \APACrefbtitle {Advances in {N}eural {I}nformation {P}rocessing
  {S}ystems} {Advances in {N}eural {I}nformation {P}rocessing {S}ystems}\
  (\BPGS\ 2672--2680).
\PrintBackRefs{\CurrentBib}

\bibitem [\protect \citeauthoryear {%
Kingma%
\ \BBA {} Welling%
}{%
Kingma%
\ \BBA {} Welling%
}{%
{\protect \APACyear {2013}}%
}]{%
wellingvae2013}
\APACinsertmetastar {%
wellingvae2013}%
\begin{APACrefauthors}%
Kingma, D\BPBI P.%
\BCBT {}\ \BBA {} Welling, M.%
\end{APACrefauthors}%
\unskip\
\newblock
\APACrefYearMonthDay{2013}{}{}.
\newblock
{\BBOQ}\APACrefatitle {Auto-encoding Variational {B}ayes} {Auto-encoding
  variational {B}ayes}.{\BBCQ}
\newblock
\BIn{} \APACrefbtitle {{Proceedings of the 2nd International Conference on
  Learning Representation (ICLR)}.} {{Proceedings of the 2nd International
  Conference on Learning Representation (ICLR)}.}
\PrintBackRefs{\CurrentBib}

\bibitem [\protect \citeauthoryear {%
Krizhevsky%
, Sutskever%
\BCBL {}\ \BBA {} Hinton%
}{%
Krizhevsky%
\ \protect \BOthers {.}}{%
{\protect \APACyear {2012}}%
}]{%
krizhevsky2012imagenet}
\APACinsertmetastar {%
krizhevsky2012imagenet}%
\begin{APACrefauthors}%
Krizhevsky, A.%
, Sutskever, I.%
\BCBL {}\ \BBA {} Hinton, G\BPBI E.%
\end{APACrefauthors}%
\unskip\
\newblock
\APACrefYearMonthDay{2012}{}{}.
\newblock
{\BBOQ}\APACrefatitle {Imagenet classification with deep convolutional neural
  networks} {Imagenet classification with deep convolutional neural
  networks}.{\BBCQ}
\newblock
\BIn{} \APACrefbtitle {Advances in {N}eural {I}nformation {P}rocessing
  {S}ystems} {Advances in {N}eural {I}nformation {P}rocessing {S}ystems}\
  (\BPGS\ 1097--1105).
\PrintBackRefs{\CurrentBib}

\bibitem [\protect \citeauthoryear {%
Lake%
, Zaremba%
, Fergus%
\BCBL {}\ \BBA {} Gureckis%
}{%
Lake%
\ \protect \BOthers {.}}{%
{\protect \APACyear {2015}}%
}]{%
lake2015deep}
\APACinsertmetastar {%
lake2015deep}%
\begin{APACrefauthors}%
Lake, B\BPBI M.%
, Zaremba, W.%
, Fergus, R.%
\BCBL {}\ \BBA {} Gureckis, T\BPBI M.%
\end{APACrefauthors}%
\unskip\
\newblock
\APACrefYearMonthDay{2015}{}{}.
\newblock
{\BBOQ}\APACrefatitle {Deep Neural Networks Predict Category Typicality Ratings
  for Images.} {Deep neural networks predict category typicality ratings for
  images.}{\BBCQ}
\newblock
\BIn{} \APACrefbtitle {Proceedings of the 37th {A}nnual {C}onference of the
  {C}ognitive {S}cience {S}ociety.} {Proceedings of the 37th {A}nnual
  {C}onference of the {C}ognitive {S}cience {S}ociety.}
\PrintBackRefs{\CurrentBib}

\bibitem [\protect \citeauthoryear {%
LeCun%
, Bengio%
\BCBL {}\ \BBA {} Hinton%
}{%
LeCun%
\ \protect \BOthers {.}}{%
{\protect \APACyear {2015}}%
}]{%
lecun2015deep}
\APACinsertmetastar {%
lecun2015deep}%
\begin{APACrefauthors}%
LeCun, Y.%
, Bengio, Y.%
\BCBL {}\ \BBA {} Hinton, G.%
\end{APACrefauthors}%
\unskip\
\newblock
\APACrefYearMonthDay{2015}{}{}.
\newblock
{\BBOQ}\APACrefatitle {Deep learning} {Deep learning}.{\BBCQ}
\newblock
\APACjournalVolNumPages{Nature}{521}{7553}{436--444}.
\PrintBackRefs{\CurrentBib}

\bibitem [\protect \citeauthoryear {%
LeCun%
\ \protect \BOthers {.}}{%
LeCun%
\ \protect \BOthers {.}}{%
{\protect \APACyear {1989}}%
}]{%
lecun1989backpropagation}
\APACinsertmetastar {%
lecun1989backpropagation}%
\begin{APACrefauthors}%
LeCun, Y.%
, Boser, B.%
, Denker, J\BPBI S.%
, Henderson, D.%
, Howard, R\BPBI E.%
, Hubbard, W.%
\BCBL {}\ \BBA {} Jackel, L\BPBI D.%
\end{APACrefauthors}%
\unskip\
\newblock
\APACrefYearMonthDay{1989}{}{}.
\newblock
{\BBOQ}\APACrefatitle {Backpropagation applied to handwritten zip code
  recognition} {Backpropagation applied to handwritten zip code
  recognition}.{\BBCQ}
\newblock
\APACjournalVolNumPages{Neural {C}omputation}{1}{4}{541--551}.
\PrintBackRefs{\CurrentBib}

\bibitem [\protect \citeauthoryear {%
Martin%
, Griffiths%
\BCBL {}\ \BBA {} Sanborn%
}{%
Martin%
\ \protect \BOthers {.}}{%
{\protect \APACyear {2012}}%
}]{%
martin2012testing}
\APACinsertmetastar {%
martin2012testing}%
\begin{APACrefauthors}%
Martin, J\BPBI B.%
, Griffiths, T\BPBI L.%
\BCBL {}\ \BBA {} Sanborn, A\BPBI N.%
\end{APACrefauthors}%
\unskip\
\newblock
\APACrefYearMonthDay{2012}{}{}.
\newblock
{\BBOQ}\APACrefatitle {Testing the efficiency of {M}arkov {C}hain {M}onte
  {C}arlo with people using facial affect categories} {Testing the efficiency
  of {M}arkov {C}hain {M}onte {C}arlo with people using facial affect
  categories}.{\BBCQ}
\newblock
\APACjournalVolNumPages{Cognitive {S}cience}{36}{1}{150--162}.
\PrintBackRefs{\CurrentBib}

\bibitem [\protect \citeauthoryear {%
Nosofsky%
, Sanders%
, Meagher%
\BCBL {}\ \BBA {} Douglas%
}{%
Nosofsky%
\ \protect \BOthers {.}}{%
{\protect \APACyear {2017}}%
}]{%
nosofsky2017toward}
\APACinsertmetastar {%
nosofsky2017toward}%
\begin{APACrefauthors}%
Nosofsky, R\BPBI M.%
, Sanders, C\BPBI A.%
, Meagher, B\BPBI J.%
\BCBL {}\ \BBA {} Douglas, B\BPBI J.%
\end{APACrefauthors}%
\unskip\
\newblock
\APACrefYearMonthDay{2017}{}{}.
\newblock
{\BBOQ}\APACrefatitle {Toward the development of a feature-space representation
  for a complex natural category domain} {Toward the development of a
  feature-space representation for a complex natural category domain}.{\BBCQ}
\newblock
\APACjournalVolNumPages{Behavior {R}esearch {M}ethods}{}{}{1--27}.
\PrintBackRefs{\CurrentBib}

\bibitem [\protect \citeauthoryear {%
Peterson%
, Abbott%
\BCBL {}\ \BBA {} Griffiths%
}{%
Peterson%
\ \protect \BOthers {.}}{%
{\protect \APACyear {2016}}%
}]{%
peterson2016adapting}
\APACinsertmetastar {%
peterson2016adapting}%
\begin{APACrefauthors}%
Peterson, J.%
, Abbott, J.%
\BCBL {}\ \BBA {} Griffiths, T.%
\end{APACrefauthors}%
\unskip\
\newblock
\APACrefYearMonthDay{2016}{}{}.
\newblock
{\BBOQ}\APACrefatitle {Adapting Deep Network Features to Capture Psychological
  Representations} {Adapting deep network features to capture psychological
  representations}.{\BBCQ}
\newblock
\BIn{} \APACrefbtitle {Proceedings of the 38th {A}nnual {C}onference of the
  {C}ognitive {S}cience {S}ociety.} {Proceedings of the 38th {A}nnual
  {C}onference of the {C}ognitive {S}cience {S}ociety.}
\PrintBackRefs{\CurrentBib}

\bibitem [\protect \citeauthoryear {%
Radford%
, Metz%
\BCBL {}\ \BBA {} Chintala%
}{%
Radford%
\ \protect \BOthers {.}}{%
{\protect \APACyear {2015}}%
}]{%
radford2015unsupervised}
\APACinsertmetastar {%
radford2015unsupervised}%
\begin{APACrefauthors}%
Radford, A.%
, Metz, L.%
\BCBL {}\ \BBA {} Chintala, S.%
\end{APACrefauthors}%
\unskip\
\newblock
\APACrefYearMonthDay{2015}{}{}.
\newblock
{\BBOQ}\APACrefatitle {Unsupervised representation learning with deep
  convolutional generative adversarial networks} {Unsupervised representation
  learning with deep convolutional generative adversarial networks}.{\BBCQ}
\newblock
\APACjournalVolNumPages{arXiv preprint arXiv:1511.06434}{}{}{}.
\PrintBackRefs{\CurrentBib}

\bibitem [\protect \citeauthoryear {%
Sanborn%
\ \BBA {} Griffiths%
}{%
Sanborn%
\ \BBA {} Griffiths%
}{%
{\protect \APACyear {2007}}%
}]{%
sanborn2007markov}
\APACinsertmetastar {%
sanborn2007markov}%
\begin{APACrefauthors}%
Sanborn, A.%
\BCBT {}\ \BBA {} Griffiths, T\BPBI L.%
\end{APACrefauthors}%
\unskip\
\newblock
\APACrefYearMonthDay{2007}{}{}.
\newblock
{\BBOQ}\APACrefatitle {Markov {C}hain {M}onte {C}arlo with people.} {Markov
  {C}hain {M}onte {C}arlo with people.}{\BBCQ}
\newblock
\BIn{} \APACrefbtitle {Advances in {N}eural {I}nformation {P}rocessing
  {S}ystems} {Advances in {N}eural {I}nformation {P}rocessing {S}ystems}\
  (\BPGS\ 1265--1272).
\PrintBackRefs{\CurrentBib}

\bibitem [\protect \citeauthoryear {%
Vondrick%
, Pirsiavash%
, Oliva%
\BCBL {}\ \BBA {} Torralba%
}{%
Vondrick%
\ \protect \BOthers {.}}{%
{\protect \APACyear {2015}}%
}]{%
vondrick2015learning}
\APACinsertmetastar {%
vondrick2015learning}%
\begin{APACrefauthors}%
Vondrick, C.%
, Pirsiavash, H.%
, Oliva, A.%
\BCBL {}\ \BBA {} Torralba, A.%
\end{APACrefauthors}%
\unskip\
\newblock
\APACrefYearMonthDay{2015}{}{}.
\newblock
{\BBOQ}\APACrefatitle {Learning visual biases from human imagination} {Learning
  visual biases from human imagination}.{\BBCQ}
\newblock
\BIn{} \APACrefbtitle {Advances in {N}eural {I}nformation {P}rocessing
  {S}ystems} {Advances in {N}eural {I}nformation {P}rocessing {S}ystems}\
  (\BPGS\ 289--297).
\PrintBackRefs{\CurrentBib}

\end{thebibliography}

\end{document}